# Ball Striking Algorithm for a 3 DOF Ping-Pong Playing Robot Based on Particle Swarm Optimization

Hossein Jahandideh, Mohammad Nooranidoost, Behnam Enghiad, Armin Hajimirzakhani

*Abstract*— This paper illustrates how a 3 degrees of freedom, Cartesian robot can be given the task of playing ping-pong against a human player. We present an algorithm based on particle swarm optimization for the robot to calculate when and how to hit an approaching ball. Simulation results are shown to depict the effectiveness of our approach. Although emphasis is placed on sending the ball to a desired point on the ping pong table, it is shown that our method may be adjusted to meet the requirements of a variety of ball hitting strategies.

## I. Introduction

Simplicity has been a priority and concentration of many robot designers during the past years. Design and implementation of Ping-pong playing robots began in the 1980's; however, until today none of the designed robots have the simplicity and low cost of a 3 degrees of freedom (DOF) robot with only translational joints and one standard ping-pong racket. In this paper, we show that such a simple robot is capable of sending a flying ball to any position on the opponent's side of the table by striking the ball correctly. We propose an algorithm to determine the robot's striking plan, though this algorithm is not limited to Cartesian robots.

One of the first ping-pong playing robots built was described in [1], where a PUMA 260 was used to play ping-pong. In [1] and most following work, the strike planning task is accomplished by the same expert controller used for controlling the robot arm. In this paper, we separate the control task from the strike planning task and consider the physical limitations of the robot performance in our strike planning algorithm.

A visual control strategy to track and intercept objects in a 3-dimensional environment was presented in [2]; this strategy was applied to a Cartesian ping-pong robot. In [2], a spherical bat was used in order for the robot to be capable of sending the ball in a desired direction. In this paper however, our objective is to use a standard ping-pong racket, rather than a modified racket or multiple rackets.

Another ping-pong player prototype was described in [3]. The robot in [3] uses two rackets on a mutual 2-DOF structure and each bat has 3 additional DOF. The objective of the robot is to send the ball to a desired point on the table. In this paper, although we emphasize on the objective of sending the ball to a desired point, unlike [3] we consider the aerodynamic model of the flight of the ball, and thus our algorithm may be adjusted to alternative objectives (as described in following sections).

The rest of this paper is organized as follows: In section II, the robot is introduced and a summary of the complete ping-pong playing process of the robot is given. In section III, particle swarm optimization is explained, which is the basis of our strike planning algorithm. In section IV, the ball's aerodynamic and rebound models are explained, which lead to our strike planning algorithm in section V. Simulation results are shown in section VI. Finally, section VII concludes the paper.

## II. Process Summary

The ping-pong playing Cartesian robot is shown in figure 1. It consists of three prismatic joints, designed such that the racket can reach any point on the robot's side of the table with a height of up to 76 cm above the table. It can also reach beyond the edges of the table by approximately 25cm.

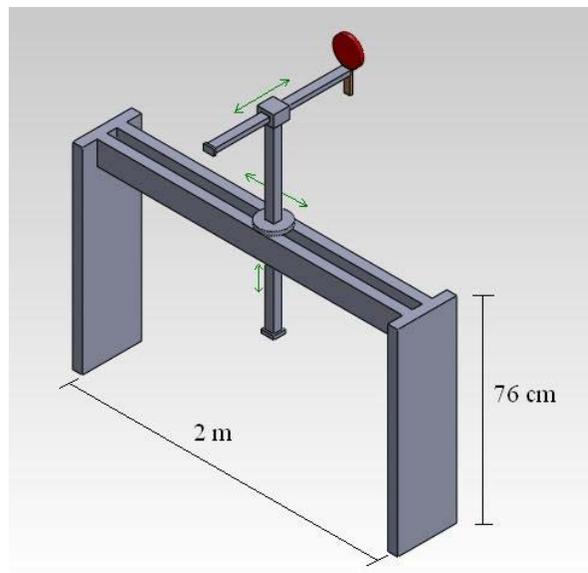

Figure 1. The Cartesian robot ping-pong player

Hossein Jahandideh is a student in the Electrical Engineering department at Sharif University of Technology, Tehran, Iran (2008-2013). hs.jahan@gmail.com

Mohammad Nooranidoost is a student in the Aerospace Engineering department at Sharif University of Technology, Tehran, Iran (2008-2013). m.nooranidoost@gmail.com

Behnam Enghiad is a student in the Chemical Engineering department at Sharif Uniersity of Technology, Tehran, Iran (2008-2013). behi08@gmail.com

Armin Hajimirzakhani is a student in the Chemical Engineering department at Sharif University of Technology, Tehran, Iran (2008-2013). ellessar_armin@yahoo.com

Figure 2 shows the procedure that the robot must follow in order to successfully play ping-pong against a human player.

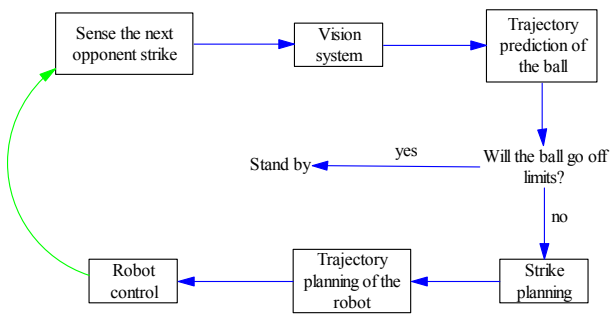

Figure 2.  Ping-pong playing process of the robot

The emphasis of this paper is on the strike planning step. The other steps have been well attended to in the literature, brief summaries of which will be given in this section.

### A. Sense the Next Opponent Strike

Once it is sensed that the opponent has made his next movement, the imaging and calculations begin immediately. This step is usually integrated into the vision system.

### B. Vision System

A vision system is required to detect the ball and calculate its position, velocity, and spin to use in the trajectory prediction step. A summary of different vision hardware that can be used is given in [4]. [4] gives a complete description of the vision system design, dividing it into sub-steps as:
B.1: Thresholding and ball detection
B.2: Camera calibration and 3-D imaging
B.3: Motion estimation
Other examples of research on vision systems for ping-pong playing robots can be found in [5-7].

### C. Trajectory Prediction of the Ball

After finding the position, velocity, and spin of the ball in one point in time, the trajectory of the ball flight can be predicted by applying physical rules to derive an aerodynamic model for the ball flight. Research on trajectory prediction of the ball can be found in [8]. The rebound model of the ball hitting the ping-pong table [5] must be considered in the trajectory prediction process. In our strike planning algorithm, the aerodynamic model of the ball flight must be considered, thus we will soon attend to this subject.

### D. Strike Planning

Having the trajectory of the ball as a function of time, the robot can choose when and how to hit the ball. The robot has four variables to set for the strike; the striking time and the x-y-z components of the velocity of the racket during the strike.

### E. Trajectory Planning of the Robot

The link variables of the robot must follow a trajectory in order to accomplish the planned strike. Methods of trajectory planning for a robot to attain specified objectives are well covered in the literature. Particularly, [9] explains trajectory planning for a robot ping-pong player.

The objective of the trajectory planning procedure is that the center of the racket must reach the planned striking point exactly at the planned striking time. Each velocity component of the racket must match its respective planned velocity at the striking time.

### F. Robot Control

Having found the trajectory that the robot must follow, all that is left is to control the robot to follow the desired trajectory. Various robot control methods are described in the *Robot Dynamics and Control* textbook [10].

## III. PARTICLE SWARM OPTIMIZATION

Particle Swarm Optimization (PSO) is a swarm intelligence optimization algorithm inspired by simulating bird flocking or fish schooling. PSO was first introduced by Kennedy and Eberhart in [11]. The mathematical analyses behind PSO were explained by Clerk and Kennedy in [12].

Let $f : \Re^n \rightarrow \Re$ be the function to be optimized. Without loss of generality, we'll take our objective to be minimization.

Objective:     minimize $f(x)$
               subject to:   $x \epsilon \chi$

The constraint $x \epsilon \chi$ can be efficiently merged with the function $f(x)$ [13]. PSO algorithm uses a swarm of k particles as agents to search for the optimal solution in an n-dimensional space. The starting position of a particle is randomly set within the range of possible solutions to the problem. The range is determined based on an intuitive guess of the maximum and minimum possible values of each component of x, but doesn't need be accurate. Each particle analyzes the function value ($f(p)$) of its current position (p), and has a memory of its own best experience (Pbest), which is compared to p in each iteration, and is replaced by p if $f(p) < f(Pbest)$. Aside from its own best experience, each particle has knowledge of the best experience achieved by the entire swarm (the global best experience denoted by Gbest). Based on the data each agent has, its movement in the *i*-th iteration is determined by the following formula:

$$V_i = w_i V_{i-1} + C_1 r_1 (Pbest - p_{i-1}) + C_2 r_2 (Gbest - p_{i-1}) \quad (1)$$

where $V_i$, $P_i$, Pbest, and Gbest are n-vectors (or similar objects, such as matrices with n components), $r_1$ and $r_2$ are random numbers between 0 and 1, re-generated at each iteration. $C_1$ and $C_2$ are constant positive numbers, $C_1$ is the *cognitive learning rate* and $C_2$ is the *social learning rate*. $w_i$ is the *inertia weight*, the importance of which is

comprehensively discussed in [14]. The new position of each particle at the *i*-th iteration is updated by:

$$p_i = p_{i-1} + V_i \quad (2)$$

After certain conditions are met, the iterations stop and the Gbest at the latest iteration is taken as the optimal solution to the problem. In this paper, we let the PSO algorithm end when the number of iterations reaches a certain number.

## IV. FLIGHT AND REBOUND MODELS

### A. Flight Model

Velocities of a flying ball are affected by gravity, viscous friction, and the Magnus force. The Magnus effect is caused by the rotational velocity (spin) of the ball. In absence of the Magnus effect, the velocity of the ball is modeled as [8]:

$$\dot{v}(t) = \begin{bmatrix} -k_d|v| & 0 & 0 \\ 0 & -k_d|v| & 0 \\ 0 & 0 & -k_d|v| \end{bmatrix} \begin{bmatrix} v_x(t) \\ v_y(t) \\ v_z(t) \end{bmatrix} + \begin{bmatrix} 0 \\ 0 \\ -g \end{bmatrix} \quad (3)$$

where $k_d$ is a constant usually obtained empirically, and g is the acceleration due to gravity. The viscous friction (air resistance) is proportional to the square of the velocity. However, we found through experiments that for the speed range of the ball in a ping-pong game, the velocity of the ball halves in each second. Thus we may replace (3) by:

$$\dot{v}(t) = \begin{bmatrix} -K_v & 0 & 0 \\ 0 & -K_v & 0 \\ 0 & 0 & -K_v \end{bmatrix} \begin{bmatrix} v_x(t) \\ v_y(t) \\ v_z(t) \end{bmatrix} + \begin{bmatrix} 0 \\ 0 \\ -g \end{bmatrix} \quad (4)$$

where $K_v$ is about 0.7. Using (4) instead of (3), allows us to model the flight with a linear time invariant (LTI) system, which greatly increases the calculation speed. As we will soon discuss, lowering the calculation time is crucial to our strike planning algorithm.

By including the effect of Magnus force, (4) turns into the following form [8]:

$$\dot{v}(t) = \begin{bmatrix} -K_v & -k_m\omega_z & k_m\omega_y \\ k_m\omega_z & -K_v & -k_m\omega_x \\ -k_m\omega_y & k_m\omega_x & -K_v \end{bmatrix} v(t) + \begin{bmatrix} 0 \\ 0 \\ -g \end{bmatrix} \quad (5)$$

where $\omega_x$, $\omega_y$, and $\omega_z$ are the rotational velocities about the x, y, and z axes respectively, and $k_m$ is proportional to the Magnus coefficient. The LTI system under study has the following form:

$$\begin{bmatrix} \dot{v} \\ \dot{x} \end{bmatrix} = A \begin{bmatrix} v \\ x \end{bmatrix} + B \quad (6)$$

where the matrices A and B have the following form:

$$A = \begin{bmatrix} -K_v & -k_m\omega_z & k_m\omega_y & 0 & 0 & 0 \\ k_m\omega_z & -K_v & -k_m\omega_x & 0 & 0 & 0 \\ -k_m\omega_y & k_m\omega_x & -K_v & 0 & 0 & 0 \\ 1 & 0 & 0 & 0 & 0 & 0 \\ 0 & 1 & 0 & 0 & 0 & 0 \\ 0 & 0 & 1 & 0 & 0 & 0 \end{bmatrix} \quad (7)$$

$$B = \begin{bmatrix} 0 & 0 & -g & 0 & 0 & 0 \end{bmatrix}^T$$

### B. Rebound Model

We wish to predict the trajectory of the ball after being struck by the robot's racket. We must first note that in this design, the robot's racket is always normal to the y-axis. The coordinate system is shown in figure 3. The origin of the coordinate system is chosen as the center of the table. In figure 3, the side of the table showing the racket is the robot's side of the table.

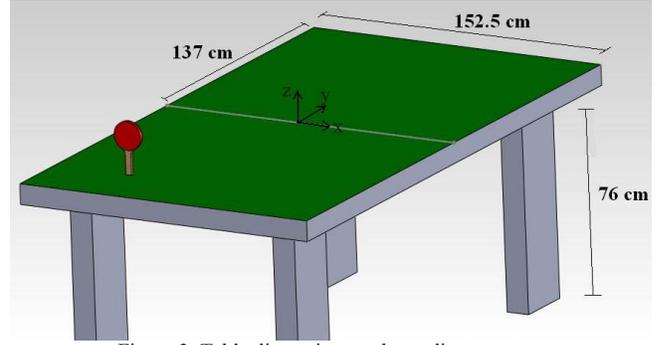

Figure 3. Table dimensions and coordinate system

In deriving the rebound model, we will first neglect the rotational velocity of the ball, and then introduce the rotational velocity to complete the model. We assume that the mass of the racket and third link of the robot is infinite with respect to the mass of the ball. We assume that the velocity of the racket is constant during the relatively small time of contact. The reference for all dynamical rules that are used in this sub-section is [15].

The y-component of the relative velocity of the ball after rebound is proportional to the relative velocity before contact.

$$v_{iy} = v_{iyb} - v_{yr} \quad (8)$$

$$v_{fy} = v_{fyb} - v_{yr} \quad (9)$$

$$v_{fy} = v_{iy} e \quad (10)$$

$$\Delta v_{yb} = v_{fy} - v_{iy} \quad (11)$$

where $v_{iab}$ and $v_{fab}$ denote the a-components of the initial and final velocity of the ball, respectively; $v_{ar}$ denotes the a-component of the velocity of the racket; $\Delta v_{ab}$ denotes the change in the a-component of the velocity of the ball (in the above definitions, a represents x, y, z or combinations of them); and *e* is the coefficient of restitution for the ball/racket impact. The coefficient *e* must be determined by the user through experiment before using this method. In this paper, we have taken $e$ to be 0.8.

In order to obtain the change in the velocity of the ball parallel to the racket, the average normal force during contact must be considered ($N_{avr}$).

$$N_{avr} = \frac{m \Delta v_{yb}}{\Delta t} \quad (12)$$

The numerator of the right side of (12) is the change in the momentum of the ball (m is the mass of the ball), and $\Delta t$ is the duration of the contact (the time it takes for the momentum change to occur).

The change in the parallel velocity of the ball is obtained by the following formulae:

$$v_{xz} = v_{xzb} - v_{xzr} \quad (13)$$

$$\Delta v_{xzb} = -\frac{N_{avr} \Delta t \mu_k}{m} \frac{v_{xz}}{|v_{xz}|} \quad (14)$$

where $\mu_k$ is the coefficient of kinetic friction, taken to be 0.2 in this paper. From (12) and (14), and by separating the x-component from the z-component of the velocities, we conclude that:

$$\Delta v_{xb} = -\frac{\Delta v_{yb} \mu_k v_{ix}}{\sqrt{v_{ix}^2 + v_{iz}^2}} \quad (15)$$

$$\Delta v_{zb} = -\frac{\Delta v_{yb} \mu_k v_{zx}}{\sqrt{v_{ix}^2 + v_{iz}^2}} \quad (16)$$

Introducing the rotational velocities into (15) and (16), they become [8]:

$$\Delta v_{xb} = -\frac{\Delta v_{yb} \mu_k (v_{ix} + r\omega_{iz})}{\sqrt{(v_{ix} + r\omega_{iz})^2 + (v_{iz} - r\omega_{ix})^2}} \quad (17)$$

$$\Delta v_{zb} = -\frac{\Delta v_{yb} \mu_k (v_{iz} - r\omega_{ix})}{\sqrt{(v_{ix} + r\omega_{iz})^2 + (v_{iz} - r\omega_{ix})^2}} \quad (18)$$

where r is the radius of the ball (taken as 2cm) and $\omega_{ia}$ is the rotational velocity of the ball about the a-axis prior to the impact. The sign of $\omega_{ia}$ is determined according to right hand rule.

The change in the rotational velocities is given by [8]:

$$\Delta \omega_x = -\frac{2}{3r} \Delta v_{zb} \quad (19)$$

$$\Delta \omega_z = \frac{2}{3r} \Delta v_{xb} \quad (20)$$

$$\Delta \omega_y = 0 \quad (21)$$

Following (17) to (21), the velocity and spin of the ball after impact are computed as follows:

$$v_{fab} = v_{iab} + \Delta v_{ab} \quad a:x,y,z$$
$$\omega_{fa} = \omega_{ia} + \Delta \omega_a \quad a:x,y,z \quad (22)$$

We conclude that through (8) to (11) and (17) to (22), the velocity and spin of the ball after impact is determined. By placing them inside the LTI flight system of (6) and using the velocity and position of the ball immediately after impact as initial conditions, the trajectory of the ball after rebound is determined.

V. STRIKE PLANNING ALGORITHM

As previously stated, the robot has 4 free variables to determine when and how to strike the ball: the striking time and the three components of velocity during impact. Let the objective of the strike be sending the ball to a specified point $(x_t, y_t)$ on the opponent's side of the table. How the target is specified depends on the strategy defined by the user. It may for example be chosen randomly on the opponent's side of the table. The trajectory of the ball after being struck by the opponent has been determined in the previous step. Take t=0 to be the time that the ball bounces on the robot's side of the table. The robot can plan the 4 free variables to attain the objective by using the following algorithm:

1- Set the free variables as parameters to be determined by the PSO algorithm. Each location that a PSO particle is in defines a striking time and 3 impact velocity components: (T, $v_{xr}$, $v_{yr}$, $v_{zr}$)

2- Define the cost function of the PSO algorithm as follows:

    a- Given the trajectory of the ball before impact, calculate $v_{ixb}$, $v_{iyb}$, $v_{izb}$ and the position of the ball at time T. (It is assumed that $\omega_i$ is previously obtained for calculation of the ball trajectory before impact).

    b- Use (6) to (11), (17) to (22), and the point of impact to obtain the trajectory of the ball after impact.

    c- Find the intersection of the post-impact trajectory of the ball with the plane z=0 and define it as: $(x_f, y_f)$

    d- The cost function is the distance between $(x_f, y_f)$ and the target $(x_t, y_t)$, or equivalently:

$$e(T, v_{xr}, v_{yr}, v_{zr}) = \|(x_f - x_t, y_f - y_t)\| \quad (23)$$

    e- The cost function is infinity if the position of the ball at time T is unreachable by the robot, or if the impact velocities of the racket are unattainable.

    f- The cost function is infinity if the ball will collide with the net in the post-impact trajectory (the net height is taken as 15 cm).

3- Apply the PSO algorithm as described in section III to find the set of arguments that minimize the cost function.

Calculation time is a very important aspect in this task. The robot must plan its free variables and reach the striking point in a limited time. The algorithm has been written in MATLAB software. We found through simulation that the

PSO algorithm usually reaches an acceptable solution within 20 iterations. The time required for 20 iterations (with a swarm population size of 10) is approximately 0.1 second, which gives the robot a high chance to plan the free variables and approach the ball on time.

We found through simulation that a solution always exists, no matter what trajectory the approaching ball has, and no matter where the target point is. As there are 4 free variables and only 2 equations to meet ($x_f=x_t$ and $y_f=y_t$), the task has redundancy. Thus, based on the strategy that the user wishes to implement, the cost function may be adjusted to meet more objectives. For example, the user may want the robot to send the ball to the desired point with a sufficiently high speed, or perhaps with a very fast spin to confuse the opponent after rebounding on the table. If (23) is used as the cost function, the algorithm will randomly converge to one solution out of infinity solutions.

## VI. SIMULATION RESULTS

A few simulated cases will be shown as examples to verify that the Cartesian robot is indeed able to play ping-pong without any extra DOF and to verify the presented strike planning algorithm. In each case, to describe the approaching ball trajectory, the velocity, spin, and position of the ball at time $t=0$ (after rebound on the table) are given, which yield the pre-impact trajectory by (6). The trajectory undergone by the ball in cases 1, 2, and 3 are shown by figures 4, 5, and 6 respectively.

The cognitive and social learning rates ($C_1$ and $C_2$ in (1)) were both set to 1.5 in our simulations, and the inertia weight ($w_i$) was chosen as constant and equal to 0.6. The swarm population was set to 10. On each case, the PSO algorithm was run for 20 iterations.

### Case 1

The cost function in this case attempts to increase the velocity with which the ball hits the table, while keeping its main objective of sending the ball to the target.

$$x_0^{(1)} = (-0.2, -0.4, 0)m$$
$$v_0^{(1)} = (0.5, -5, 4)m/s \quad (24)$$
$$\omega_0^{(1)} = (10, 10, 10)rad/s$$

$$e_1(p) = \|(x_f + 0.3, y_f - 1)\| + \frac{0.5}{\|v_{tar}\|} \quad (25)$$

where $x_0$ denotes the position of the ball at the time $t=0$, $e_1$ is the cost function used in case 1, and $v_{tar}$ is the velocity with which the ball hits the opponent's side of the table.

The PSO algorithm reached the following solution:

$$p_1^* = (T^*, v^*_{xr}, v^*_{yr}, v^*_{zr}) = (0.82, -0.25, 0.91, 1.14) \quad (26)$$

Following the planned strike, the ball is sent exactly to the target point $(x_f, y_f)=(-0.3,1)$ with an absolute velocity of 6.6 m/s.

### Case 2

In this case, the cost function is designed to increase the rotational velocities about the x and y axes in the post-impact trajectory.

$$x_0^{(2)} = (0.5, -0.1, 0)m$$
$$v_0^{(2)} = (0.7, -2, 3)m/s \quad (27)$$
$$\omega_0^{(2)} = (-6, 5, 0)rad/s$$

$$e_2(p) = \|(x_f - 0.75, y_f - 0.3)\| + \frac{1}{\sqrt{\omega_{fx}^2 + \omega_{fy}^2}} \quad (28)$$

$$p_2^* = (T^*, v^*_{xr}, v^*_{yr}, v^*_{zr}) = (1.35, -2.5, 1.09, 0.03) \quad (29)$$

### Case 3

In this case, the cost function is adjusted to increase the y-component of the ball's velocity at the time it hits the target, while lowering the z-component to an extent.

$$x_0^{(3)} = (0, -1, 0)m$$
$$v_0^{(3)} = (0, -5, 3)m/s \quad (30)$$
$$\omega_0^{(3)} = (1, 5, -5)rad/s$$

$$e_3(p) = \|(x_f + 0.2, y_f - 1)\| + 0.01|v_{tz}| + \frac{0.5}{|v_{ty}|} \quad (31)$$

where $v_{ty}$ and $v_{tz}$ are the y and z components of the velocity of the ball at the time of collision with the table.

$$p_3^* = (T^*, v^*_{xr}, v^*_{yr}, v^*_{zr}) = (0.51, -1.61, 1.02, 3.43) \quad (32)$$

The weight coefficients in (25), (28), and (31) are chosen such that a higher priority is given to reaching the target, while the coefficients of the other objectives are big enough to keep the objectives from becoming insignificant. The values of the terms representing all objectives other than the target must be kept in the same order of magnitude.

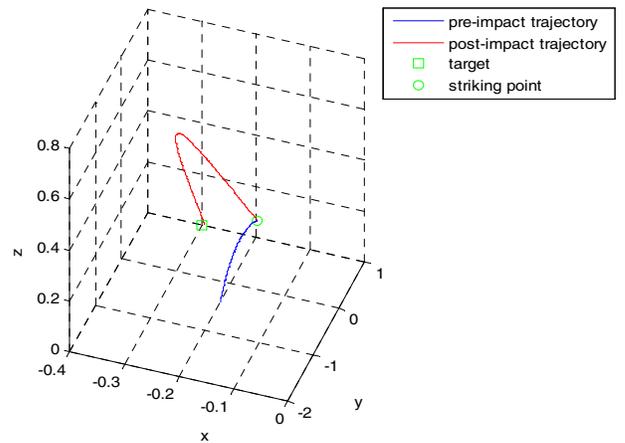

Figure 4. Ball trajectory for the first case of simulation

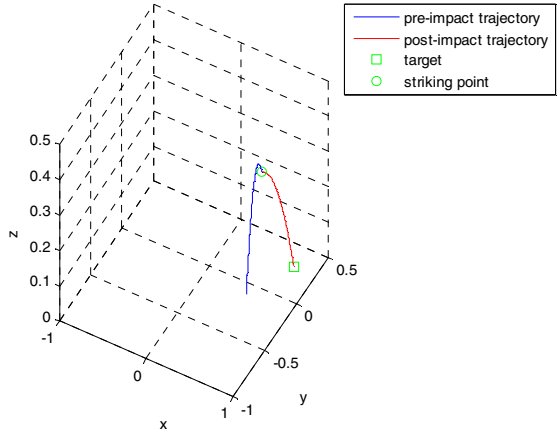

Figure 5.  Ball trajectory for the second case of simulation

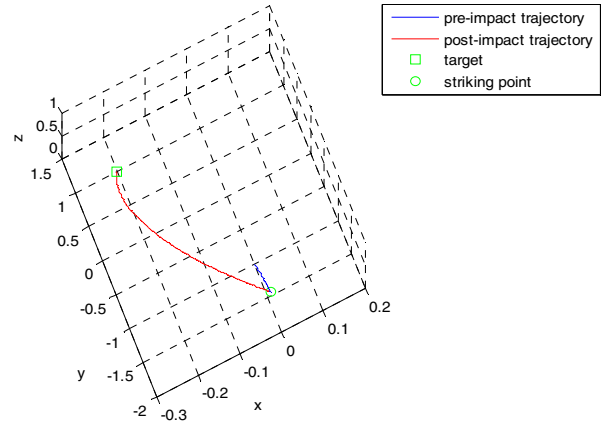

Figure 6.  Ball trajectory for the third case of simulation

A few more examples of sample simulations are shown in table 1.

TABLE I: RESULTS OF FOUR SAMPLE SIMULATIONS

| Target (x,y) | Other objective | Reached point (x,y) | Other objective accomplishment | Computation time (s) |
|---|---|---|---|---|
| (0,0.5) | Increase the maximum height of the ball | (-0.01,0.50) | Maximum height reached by the ball: 1.07 m | 0.089 |
| (-0.7,0.3) | Increase the spin about x axis | (-0.66,0.33) | Spin about the x axis: 16 rad/s | 0.087 |
| (0.3,1.37) | Decrease the maximum height of the ball | (0.3,1.37) | Maximum height reached by the ball after rebound: 18 cm | 0.072 |
| (0,0) | (invalid target; on the net) | (0,0.12) | (reached the closest reachable feasible target) | 0.063 |

## VII.  CONCLUSION AND FUTURE WORK

It has been shown that a robot as simple and low cost as a Cartesian robot holding a standard racket can be programmed to play ping-pong against a human player. A PSO-based algorithm was proposed to determine when and how to hit the ball. This algorithm, aside from having a near perfect success rate at throwing the ball to a specified target, can also be adjusted to follow various strategies, such as the ball reaching the target with maximum speed, or with maximum spin, etc.

In future work, we will build the robot and have it play against humans. However, before going forward to the implementation stage, we plan to find a model to predict the error caused by the simplifications made (such as the assumption of an LTI system in the flight model). We will compensate for the predicted errors by using feed-forward in the robot control stage. This step is necessary because if the robot reaches the striking point too late or too soon, it may either miss the ball or send it to an undesired direction.